\documentclass[letterpaper, 10 pt, conference]{ieeeconf}  

\IEEEoverridecommandlockouts                              

\usepackage{graphics} 
\usepackage{epsfig} 
\usepackage{mathptmx} 
\usepackage{times} 
\usepackage{amsmath} 
\usepackage{amssymb}  
\usepackage{slashbox}
\usepackage{makecell}
\usepackage{cite}

\title{\LARGE \bf
Feature Analysis and Selection for Training an End-to-End Autonomous Vehicle Controller Using the Deep Learning Approach
}

\author{Shun Yang$^{1,2}$, Wenshuo Wang$^{1,3}$, Chang Liu$^{1}$, Kevin Deng$^{2}$ and J. Karl Hedrick$^{1}$
\thanks{*This project is partly supported by National Science Foundation of China under Grant U1564211 and National Key Research and Development Program of China under Grant 2016YFB0100904.}
\thanks{$^{1}$Authors are with the Department of Mechanical Engineering, University of California, Berkeley, USA
        {\tt\small changliu@berkeley.edu, khedrick@me.berkeley.edu }}%
\thanks{$^{2}$Authors are with College of Automotive Engineering,
Jilin University, China
        {\tt\small yangshun628@gmail.com, kdeng@jlu.edu.cn}}%
\thanks{$^{3}$Authors are with School of Mechanical Engineering,
Beijing institute of technology, China
        {\tt\small wwsbit@gmail.com}}%
}

\begin{document}

\maketitle
\thispagestyle{empty}
\pagestyle{empty}

\begin{abstract}
Deep learning-based approaches have been widely used for training controllers for autonomous vehicles due to their powerful ability to approximate nonlinear functions or policies. 
However, the training process usually requires large labeled data sets and takes a lot of time. 
In this paper, we analyze the influences of features on the performance of controllers trained using the convolutional neural networks (CNNs), which gives a guideline of feature selection to reduce computation cost. 
We collect a large set of data using The Open Racing Car Simulator (TORCS) and classify the image features into three categories (sky-related, roadside-related, and road-related features).
We then design two experimental frameworks to investigate the importance of each single feature for training a CNN controller.
The first framework uses the training data with all three features included to train a controller, which is then tested with data that has one feature removed to evaluate the feature's effects. 
The second framework is trained with the data that has one feature excluded, while all three features are included in the test data. 
Different driving scenarios are selected to test and analyze the trained controllers using the two experimental frameworks. 
The experiment results show that (1) the road-related features are indispensable for training the controller, (2) the roadside-related features are useful to improve the generalizability of the controller to scenarios with complicated roadside information, and (3) the sky-related features have limited contribution to train an end-to-end autonomous vehicle controller.

\end{abstract}

\section{Introduction}

Deep learning has drawn lots of attentions in recent few years in the realms such as classification, natural language processing, dimension reduction, object detection, and motion modeling \cite{3,4,5}, due to its powerful ability to approximate highly nonlinear functions \cite{2}. 
Convolutional neural net (CNN) is one approach to implementing deep learning and particularly suited for image recognition as it can perform a dimensional reduction of high-dimensional inputs through convolution \cite{6}. 
Therefore, CNN has been introduced to deal with problems in autonomous vehicles, such as detection and classification of pedestrians and vehicles \cite{ishak2006face,tian2015pedestrian,du2016fused}, and environment perception\cite{dp2}. 
In addition, researchers have implemented CNN for an end-to-end framework of learning an autonomous vehicle controller. 
Compared to explicit decomposition of controller design methods, such as lane marking detection, path planning, and vehicle control, training an end-to-end controller can simultaneously optimize these steps. 
For example, in \cite{lecun2005off}, robots with an end-to-end vehicle controller can detect obstacles and navigate around them in real time.
Xu, {\it et al}. \cite{xu2016end} also used the CNN to learn an end-to-end vehicle controller that can follow the curved lane accurately. 
In \cite{6},  the car with an end-to-end controller can run in traffic with/without lane markings. It also performs well in the scenarios with unclear visual guidance such as in parking lots and on unpaved roads.

However, since training CNNs needs a large amount of labeled data covering diverse scenarios, the training procedure is always computationally expensive and time-consuming. 
Researchers have utilized multiple Graphics Processing Units (GPUs) to cope with this problem\cite{simonyan2014very}. 
This, however, increases the development cost. 
In addition, the program architecture of GPUs differs significantly from the traditional central process units, which makes coding with GPUs hard to grasp.  
Li, {\it et al}. \cite{li2015convolutional} used low-resolution images to shorten the training time, but this leads to reduced training accuracy.
 
Researchers in \cite{shalev2016sample} and \cite{ohn2017all} ranked the object-level importance in images by training the neural network with semantic abstraction or human-centric annotations. 
However, it is still unclear how an end-to-end CNN controller can perform if trained by discarding less important features in images and applied to normal driving scenarios.

This paper analyzes the importance of different image features for training an end-to-end autonomous vehicle controller. 

We describe a neural net architecture and the training of a CNN-based end-to-end steering controller of autonomous vehicles. 
Then, we propose two frameworks to analyze the importance of different features for learning the CNN. 
In this paper, image features are classified into three categories and new data sets with only one of three features excluded are used for training and validation. 
The performance of learned controllers are also validated and analyzed by running a closed-loop test in simulation to control an autonomous vehicle running in different tracks.

\begin{figure*}[t]  	
	\centering
	\includegraphics[width=1\textwidth]{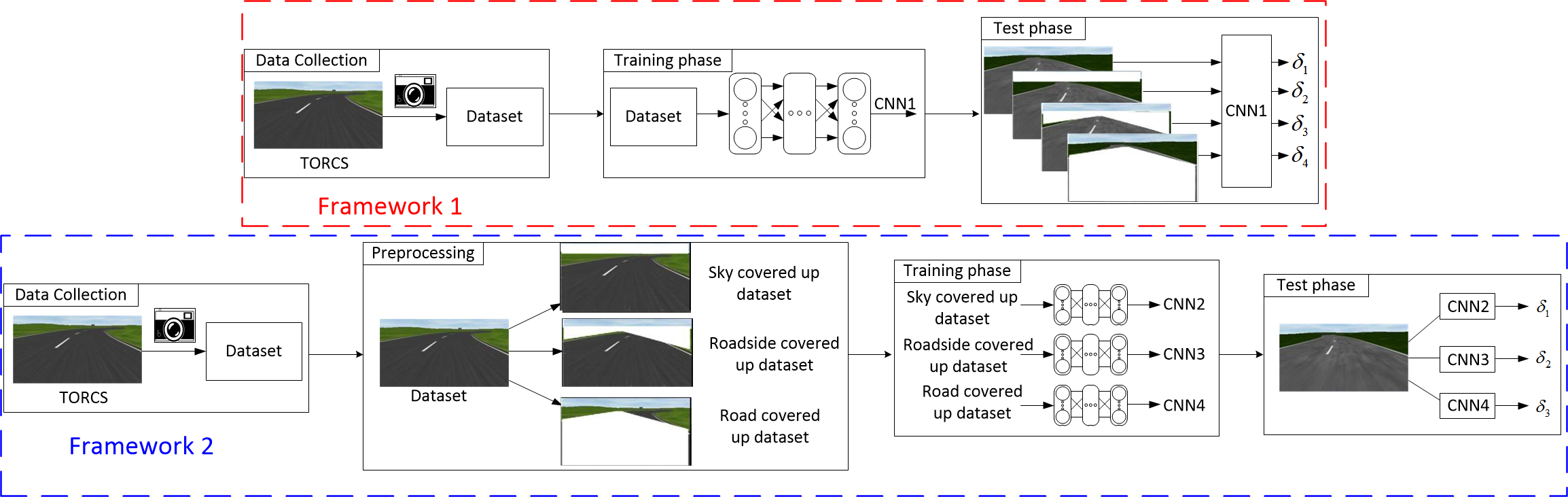}
	\caption{Two frameworks for feature analysis}
	\label{fig1}
\end{figure*}	

Section II presents the method to train end-to-end autonomous vehicle controllers and the experiment design. The learned controllers are validated in Section III.
Section IV presents the feature evaluation. 
Conclusions and future works are discussed in Section V.

\section{Methods}
We collected the images (i.e., driving scenarios) with labels (i.e., steering angle) using The Open Racing Car Simulator (TORCS), which is widely used in AI research. Then, image features of the collected data sets are grouped. Two different frameworks are proposed to train and test the CNNs, as shown in Fig. \ref{fig1}. Based on the two frameworks, we can evaluate the importance level of feature for learning a controller based on CNNs.

\subsection{Data Collection}
\begin{figure}[t]  	
	\centering
	\includegraphics[width=0.48\textwidth]{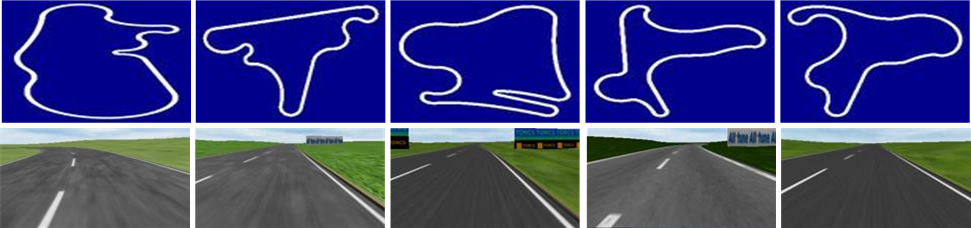}
	\caption{Examples of tracks used for training.}
	\label{fig2}
\end{figure}

To train a CNN, we need to label images by matching each scenario screenshot with a steering angle. In this paper, the labeled data is collected by a human driver driving cars in TORCS with joystick wheel. Here, the experimental scenarios include 13 different double-lane tracks without other road users. Examples of five tracks are shown in Fig. \ref{fig2}. We replace the original road surface textures in TORCS by customized asphalt textures and asphalt darkness levels so that data coherence can be guaranteed. We sample the images with 10 frames per second (FPS), because more FPS would only lead to more similar frames without providing more useful information\cite{6}. The labeled screenshots are down-sampled to 190*100 and stored in a database together with the normalized steering angle from -1 to 1. The car speed is set as a constant of 60 km/h. Then, we generate a Hierarchical Data Format 5 (HDF5) file that includes frames and steering angles. This file is the input of training process in CAFFE\cite{jia2014caffe}. CAFFE is a well-known deep learning framework developed by the Berkeley Vision and Learning center (BVLC) . Totally 33,700 training images are included in the training data set.

\subsection{Network Structure}
\begin{figure*}[t]  	
	\centering
	\includegraphics[width=1\textwidth]{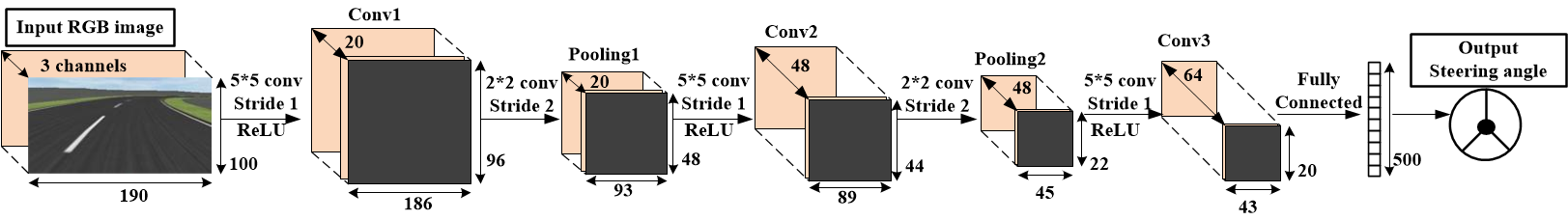}
	\caption{Net structure of the designed neural net. The net consists of three convolutional layers, two pooling layers and two fully connected layers. Dropout is used to prevent overfitting. A $190\times 100$ crop of an image (with 3 color planes) is presented as input, which is convoluted with 20 different $1^\text{st}$ layer kernels as the filter, each of size $5\times 5$, using a stride of 1 in both x and y. The resulting feature maps are then passed through a rectified linear function, pooled (max within 2 by 2 regions, using stride 2), and contrast normalized across feature maps to give 20 different $48\times 93$ element feature maps. Similar operations are repeated in Conv2 and Conv3. The fully connected layer takes features from the top convolutional layer and transfer them into vector form.
	}
	\label{fig3}
\end{figure*}

We define the net structure in CAFFE, as shown in Fig. \ref{fig3}. We use a CNN with four hidden layers, including three convolutional layers and one fully connected layer. The input data has the size 190*100*3 and we use a batch size of 100\footnote{A batch is a subset of input frames.}. The first two convolutional layers have a kernel size of 5*5. The first one has 20 feature maps as output and the second one has 48 feature maps as output. They are used to extract the features of roads. After the first two layers we use a pooling layer with 2*2 kernel to scale the frames\footnote{The pooling layer operates max operation to resize the feature map.}. The third convolutional layer has a 3*3 kernel size. The fully connected layer has 500 outputs and the last layer has the steering angle as an output. The last layer is called the Euclidean loss layer. This layer calculates the loss E, given by

$$
{\emph{E}}=\frac{1}{2N} {\sum_{k=1}^N} { \lVert \hat{y}_n-y_n \rVert}_2^{2} \eqno{(1)}$$
where $\hat{y}_n$ is the estimated value and  $y$ is the labeled value after
an iteration with the batch size $N$. 
The loss is necessary for the optimization algorithm to update the weights and bias to minimize the loss of the next iteration.

After each layer, we use a dropout to prevent the neural net from overfitting. To achieve a faster learning process, we use the rectified linear unit as an activation function. 

\subsection{Experiment Design}

\subsubsection{Framework 1}
\begin{figure}[t]  	
	\centering
	\includegraphics[width=0.42\textwidth]{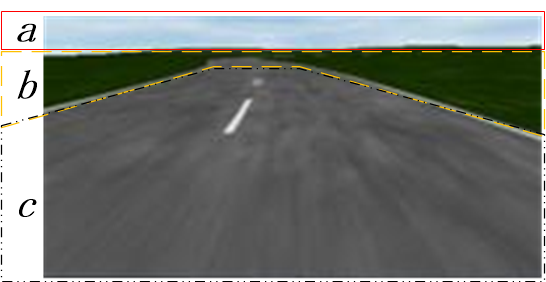}
	\caption{ Illustrations of typical feature areas.}
	\label{fig_add}
\end{figure}

In Framework1, 33,700 images are taken as a training data set to learn an end-to-end autonomous vehicle controller, denoted as CNN1. After the training converges, CNN1 would be tested using some unknown test data sets. Then, features of test data sets are manually classified into 3 categories, i.e., sky-related feature, roadside-related feature, and road-related feature.  Typical feature areas are represented in Fig. \ref{fig_add}.  Sky-related feature (region {\textbf{a}} in Fig. \ref{fig_add}) refers to the upper part of an image, often with clouds, birds or some part of skyscraper; roadside-related feature (region {\textbf{b}} in Fig. \ref{fig_add} ) denotes the middle right and left part of an image, often with grass, trees and buildings; road-related feature (region {\textbf{c}} in Fig. \ref{fig_add}) is the lower-middle part of an image, often with road of different textures. In the analysis stage, each feature will be covered up one-by-one to evaluate its importance. 

\subsubsection{Framework 2}
In Framework2, the training data set with 33,700 images is preprocessed by covering a single feature shown in Fig. \ref{fig_add} with a white polygon, and thus three new training data sets are obtained with different features covered, i.e., sky covered-up data set, roadside covered-up data set and road covered-up data set. Then three different end-to-end autonomous vehicle controllers (CNN2, CNN3, and CNN4) are also trained in CAFFE using the three different covered-up data sets, respectively. After that, three controllers are tested using the data set with the same features as the corresponding training data set.

In addition, data sets consisting of all three features are used to test controllers CNN2, CNN3, and CNN4. The importance level of each single feature can be assessed. Then, to evaluate the features' importance for training a CNN, the three controllers are used to handle the car in TORCS game. The duration time of successfully tracking is recorded and compared.

\section{Validation of Trained Controllers}
We evaluate the effectiveness of different controllers by checking the loss values for all learned controllers.
\subsection{Verification and Validation of CNN1}
\begin{figure}[t]  	
	\centering
	\includegraphics[width=0.48\textwidth]{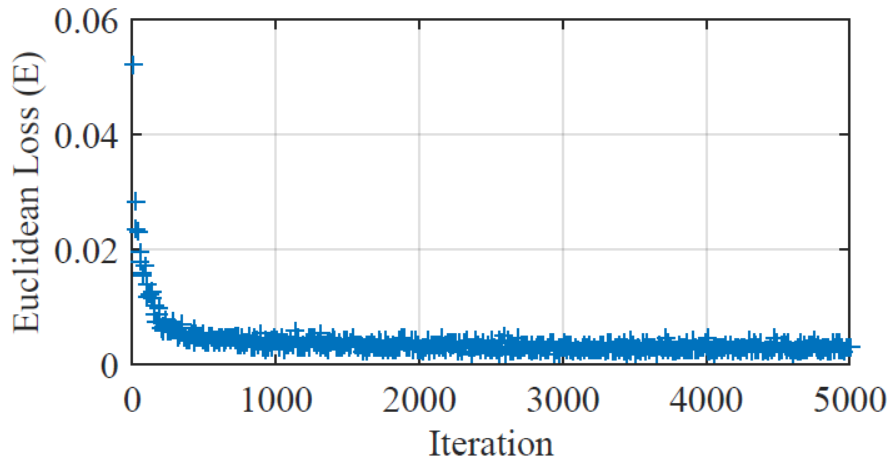}
	\caption{Euclidean loss of CNN1 in 5000 iterations.}
	\label{fig4}
\end{figure}

First, the end-to-end controller CNN1 is learned using the data set consisting of 33,700 images without any feature covered up. We train the controller on a laptop with only the CPU. The training can also be accelerated by exploiting GPUs. We evaluate the effectiveness of training by examining the loss values. As soon as the loss value converges and does not decrease anymore, the net is either trained well enough or the net structure is not suitable for the task at hand. For the case where the net is already sufficiently trained and the training process is not stopped, overfitting issues may occur even if dropouts are used. Fig. \ref{fig4} shows loss value of CNN1 over the training iterations. We note that, after the 5,000$^\text{th}$ iteration, the loss value of CNN1 does not decrease noticeably, so the training process can be stopped at this stage.

\begin{figure}[t]  	
	\centering
	\includegraphics[width=0.48\textwidth]{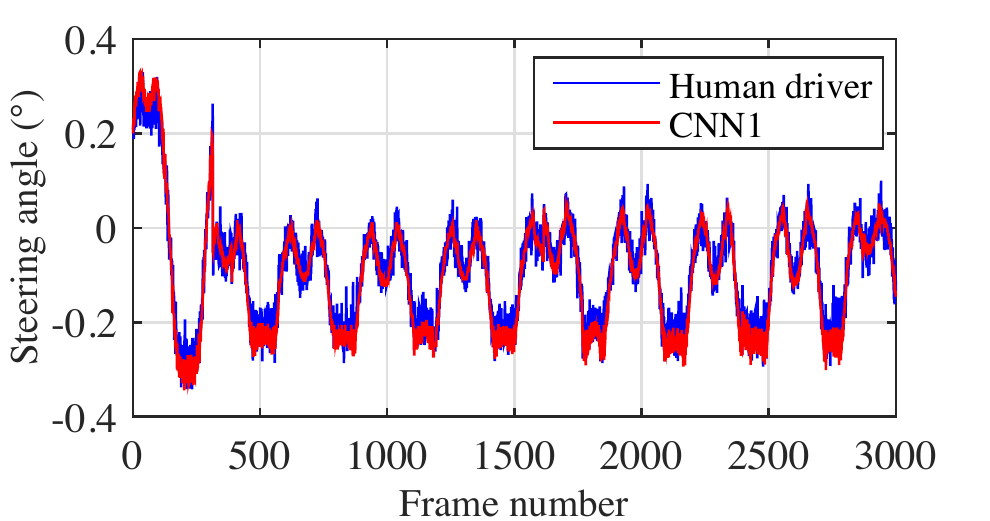}
	\caption{Comparison of steering angles from a human driver and the CNN1 for a new data set.}
	\label{fig5}
\end{figure}

The point of convergence of loss values is a good indicator for a well trained neural net, but the most important factor is how well the trained net performs on new data. Fig. \ref{fig5} compares the  steering angles from a human driver and the steering angles estimated using CNN1 for the new data set. We can see that CNN1 performs well on new data, indicating that the structure of the designed neural network is reasonable and the controller trained is effective. 

\subsection{Verification and Validation of CNN2, CNN3, and CNN4}
\begin{figure}[t]  	
	\centering
	\includegraphics[width=0.48\textwidth]{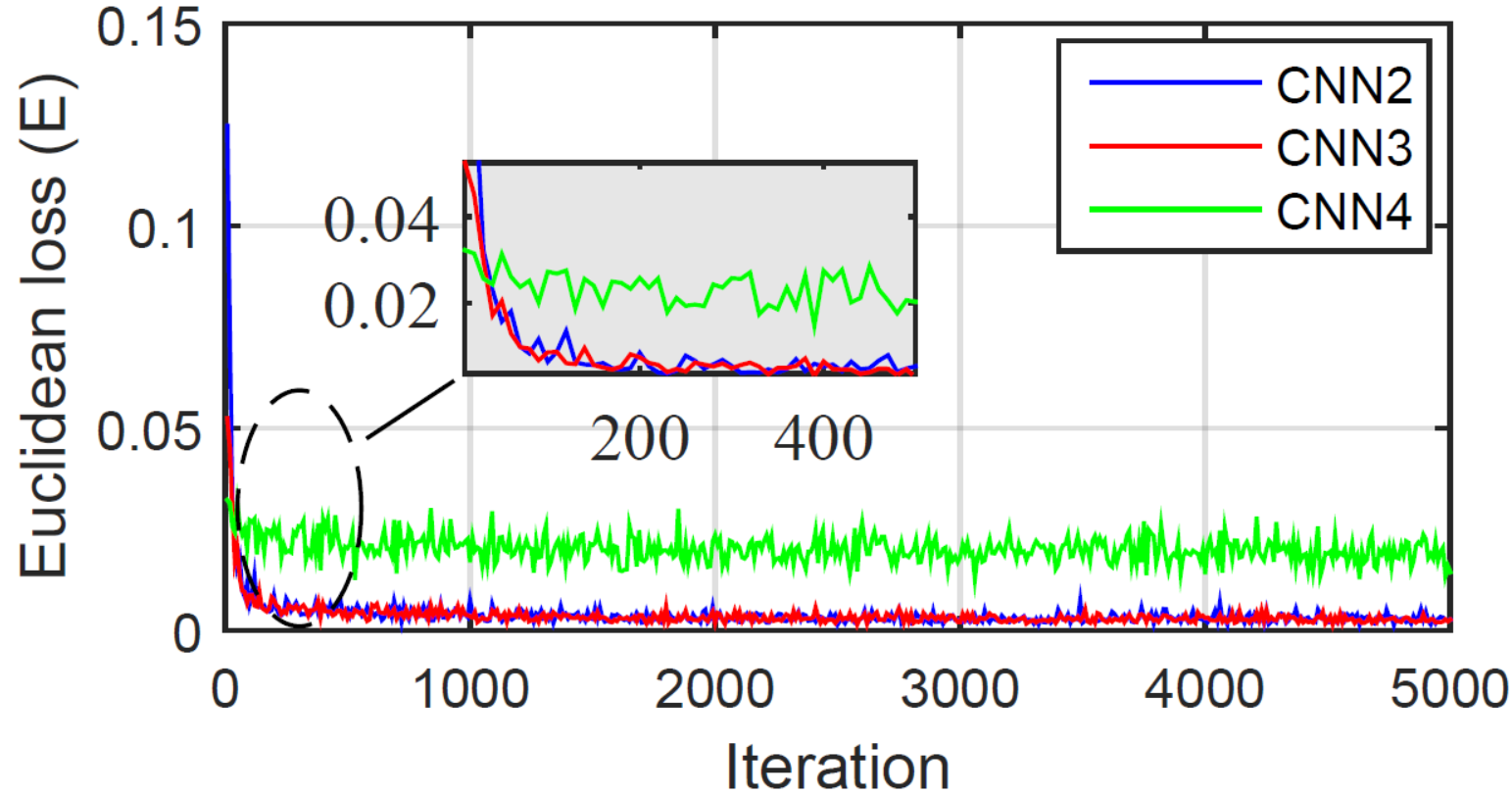}
	\caption{Euclidean loss of CNN2, CNN3, and CNN4 over 5000 iterations.}
	\label{fig6}
\end{figure}

We generate 3 new data sets from the data set by discarding different features.
Fig. \ref{fig6} shows the loss values of CNN2, CNN3, and CNN4 over the training iterations. After 5,000 iterations, the loss value does not decrease noticeably, so the training processes are stopped. We note that the end-to-end controllers trained using the data set with sky covered-up and roadside covered-up have almost similar rate of convergence, i.e., both of them decease significantly in the first 500 iterations and converge at the 1000$^{\text{th}}$ iteration. However, as to the controller trained using the data set with road covered-up, the loss value decreases to 0.02 at the very beginning and oscillate around 0.02 till the end.

In the verification process of CNN1, the net structure is proved to be good, which are also verified by the convergence processes of CNN2 and CNN3, as seen in Fig. \ref{fig6}. So the reason for the non-convergence of CNN4 would be the road feature excluded dataset. It is easy to understand that even for human beings, if driving without knowing anything about the road, it would be hard to decide how to behave next.

\begin{figure}[t]  	
	\centering
	\includegraphics[width=0.48\textwidth]{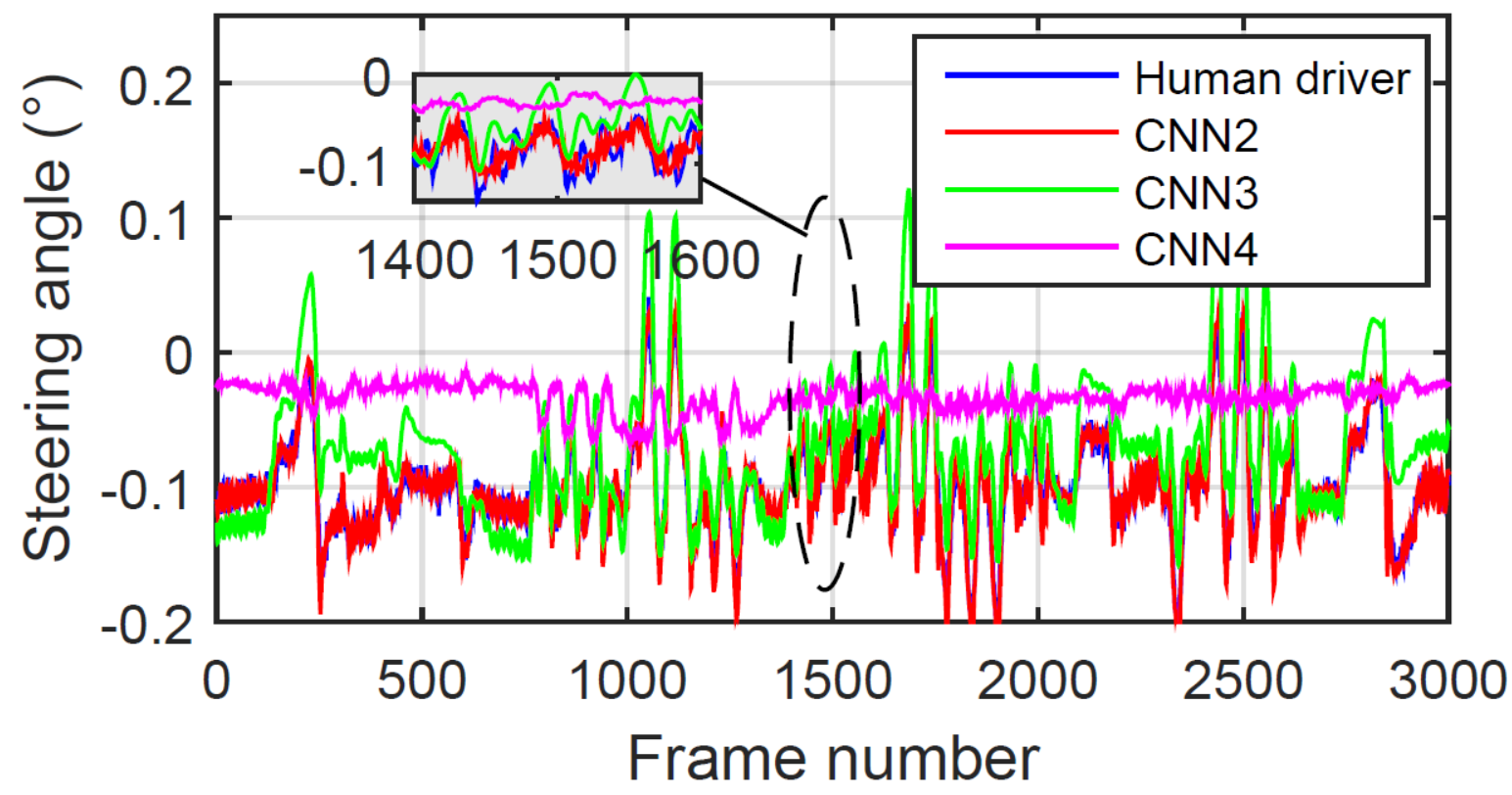}
	\caption{Comparison of steering angles from the human driver and three end-to-end controllers for  data sets with new feature covered.}
	\label{fig7}
\end{figure}

Then, we test three controllers separately using new data sets. These data sets differ from each other because each data set is with only one feature covered-up. Fig. \ref{fig7} shows test results of CNN2, CNN3, and CNN4, compared to the human driver's steering angles. The results of 1400-1600 iterations is magnified for a detailed comparison.
Table \ref{table1} shows means and standard deviations of the Euclidean loss between the predicted and labeled steering angles for CNN2, CNN3, and CNN4. We note that CNN2, trained using the sky covered-up data set, has the smallest mean loss value of $1.827*10^{-4}$. The mean loss values of other two controllers are higher than CNN2 by one magnitude. The CNN3, trained with roadside covered-up data set, has the mean loss of 0.0019 and is much better than CNN4 which is trained using data with road covered-up.

\begin{table}[h]
	\caption{Means and Standard Deviations of The Euclidean Loss Between The Scaled Predicted and Labeled Steering Angles for Test Data Sets with Feature Covered.}
	\label{table1}
	\begin{center}
		\begin{tabular*}{7.2cm}{@{\extracolsep{\fill}}lllr}
			\hline
			 & CNN2 & CNN3 & CNN4\\
			\hline
			Mean error & $1.827*10^{-4}$  &  0.0019 & 0.0056\\
			\hline
			Standard deviation & $3.362*10^{-4}$ &  0.0022 & 0.0045\\
			\hline
		\end{tabular*}
	\end{center}
\end{table}
\section{Features Evaluation}
We use  CNN2, CNN3, and CNN4 controllers to evaluate the importance level for sky-related features, roadside-related features, and road-related features, respectively. The evaluation is also carried through the proposed two frameworks.
%
%
%
%
\subsection{Framework1}
In Framework1, CNN1 is test using data set with each single feature covered-up one-by-one. This provides us a direct understanding of the relationship between controller output and importance level of features in scenarios. 

\begin{figure}[t]  	
	\centering
	\includegraphics[width=0.48\textwidth]{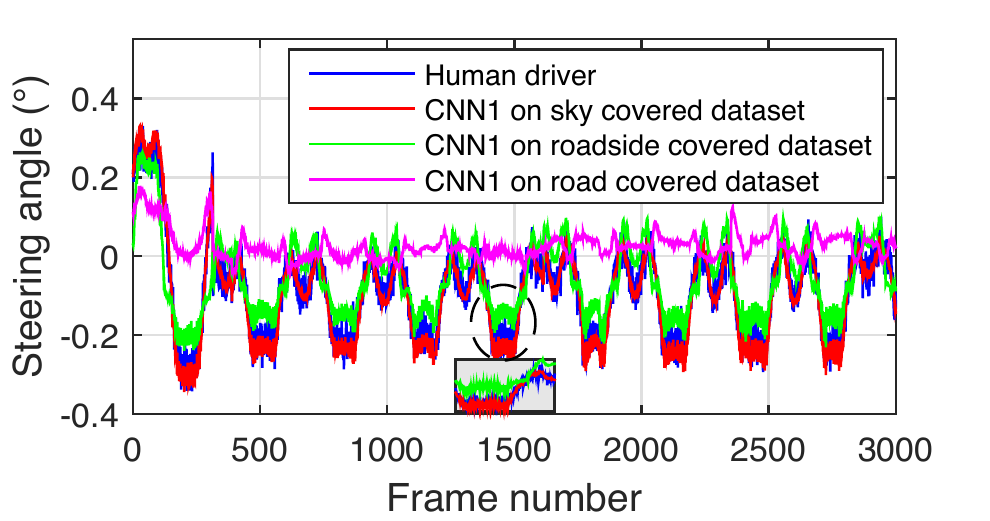}
	\caption{Comparison of the steering angle from human driver and from the end-to-end controller CNN1 for data sets with new features covered-up.}
	\label{fig8}
\end{figure}

Fig. \ref{fig8} presents the predicted steering angles from CNN1 with different features covered-up, compared with the steering angles from human drivers. From Fig. \ref{fig8} we note that the predicted steering angle using the data set with the sky-related feature covered-up can highly match with the steering angles from human drivers. This indicates that the neural network can 'drive' the vehicle well without knowing any sky-related information. The green line is the predicted steering angle using the data set with the roadside covered-up. It is little bit different from the steering angles from human drivers, but the shape of steering angle is similar to the steering angles from human drivers. We can understand this in the way that without knowing the roadside information, the CNN1 controller still works well because the road features can provide partial useful information for drivers, but to obtain an accurate control command, the roadside features are also needed.

For road-related features, the pink line (Fig. \ref{fig8}.) is really far from the human driver's control, which means that the road-related feature in the data set is the most important for learning.

\subsection{Framework2}
Framework2 evaluates the features from an opposite view, compared with Framework1. In Framework2, we test the three controllers (CNN2, CNN3, and CNN4) using the data set with all-features involved. In this way, we can evaluate the influence of different features on training an end-to-end autonomous vehicle controller.

\begin{figure}[t]  	
	\centering
	\includegraphics[width=0.48\textwidth]{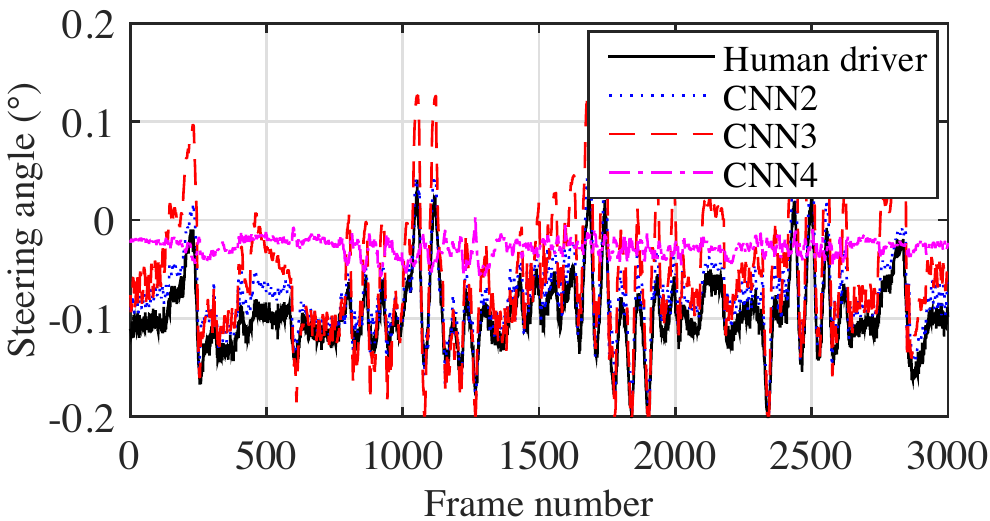}
	\caption{Comparison of the human driver and predicted steering angle from controllers CNN2, CNN3, CNN4 on unknown all-featured dataset}
	\label{fig9}
\end{figure}

Fig. \ref{fig9} compares the human driver and three end-to-end controllers for predicting steering angles on a new data set with all features included. In Fig. \ref{fig9}, none of the controllers can perform as much well as the labeled value. CNN2 behaves better than the other two and CNN4 behaves in irregular. Since CNN4 is trained using the data set without road-related features, it can be inferred that the road feature is indispensable for the end-to-end controller training. On the contrary, the sky-related features are the least important for an end-to-end autonomous controller.

Table \ref{table2} shows the means and standard deviations of the Euclidean loss between the predicted and labeled steering angles of CNN2, CNN3, and CNN4. We can see that the end-to-end controller CNN2 has the smallest mean loss value of $6.095*10^{-4}$. The mean loss values of CNN3 and CNN4 are 0.003 and 0.0062 and higher than CNN2 by one magnitude. Compared to Table \ref{table1}, we can find that the performance of CNN controllers trained with missing feature datasets would degrade in different degree, which indicates that the deviation extent and dispersion degree become larger if we apply the controller trained without some features to the test scenarios with these features included.

\begin{table}[t]
	\caption{Mean and Standard Deviation of the Euclidean Loss Between the Scaled Predicted and Labeled Steering Angles (for the Same All-Feature-Included Test Data Set).}
	\label{table2}
	\begin{center}
		\begin{tabular*}{7.2cm}{@{\extracolsep{\fill}}lllr}
			\hline
			& CNN2 & CNN3 & CNN4\\
			\hline
			Mean error & $6.095*10^{-4}$  &  0.0030 & 0.0062\\
			\hline
			Standard deviation & $6.614*10^{-4}$ &  0.0035 & 0.0050\\
			\hline
		\end{tabular*}
	\end{center}
\end{table}

\begin{figure}[t]  	
	\centering
	\includegraphics[width=0.48\textwidth]{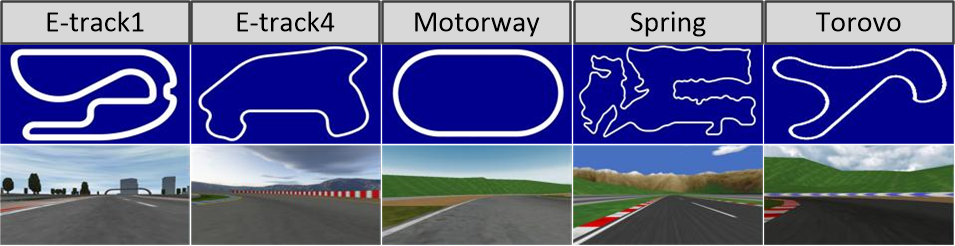}
	\caption{Illustration of  tracks and scenarios selected for real time TORCS car control}
	\label{fig10}
\end{figure}

To further analyze the influence of discarding features on the training an end-to-end autonomous controller, we then implement the three controllers (i.e., CNN2, CNN3, and CNN4) in the closed-loop driving test. Fig. \ref{fig10} shows the new tracks and scenarios for the close-loop test. In the closed-loop test, screenshot of scenarios is input to controllers to generate steering angle for the whole track control. 

Table \ref{table3} shows the lasting time of successfully tracking the lane for a TORCS car in different test scenarios. From Table \ref{table3}, we can see that the TORCS car with CNN4 drifts off the track at 15s in every scenario. In simpler scenarios such as Motorway and Spring, the three controllers (CNN1, CNN2, and CNN3) perform well and can keep a TORCS car in the track and  successfully finish the tracking test.  For the track with complicated and colored road edges (e.g., E-track1 and Torovo), all three controllers cannot keep the car in the lane for a long time because of the sudden turn and confusing features in the track, and the lasting time of successful tracking are almost similar. From the tests in E-track1, Spring and Motorway, the end-to-end controllers trained with/without the sky- and roadside-related features have a similar performance on the simple or complicated scenarios.

The most interesting scenario is E-track4. The shape of E-track4 is quite simple and not any sharp turn exists in the track. However, the roadside of E-track4 consists of different textures such as grass and sand. In this scenario, the car with controller CNN1 can finish the track, but the car with CNN2 and CNN3 can only keep in the track for 331.2s and 127.7s, respectively. Since CNN2 is trained using the data set with roadside-related features included and thus has a good generalization capability for different roadside scenarios, therefore it can run much longer than CNN3 in a track with grass and sand features mixed. From another aspect, we can conclude that the priority of features for training an end-to-end controller is ranked as road-related features, the roadside-related features, and the sky-related features.


\begin{table}[t]
	\caption{Lasting Time of Successful Tracking Lane Using Different Controllers in New Tracks.}
	\label{table3}
	\begin{center}
		\begin{tabular}{|l|*{4}{c|}}\hline
			\backslashbox{Tracks}{Controllers}
			&\makebox[3em]{CNN1}&\makebox[3em]{CNN2}&\makebox[3em]{CNN3}
			&\makebox[3em]{CNN4}\\\hline
			E-track1 &65&62.26&62.02&10.28\\\hline
			E-track4 &Finished&331.2&127.7&12.82\\\hline
			Motorway &Finished&Finished&Finished&14.2\\\hline
			Spring &444&447&445.1&12.62\\\hline
			Torovo &97.46&95.72&97.3&11.8\\\hline
		\end{tabular}
	\end{center}
\end{table}

\section{Conclusions and Future Work}
From the experiment validation and discussion above, we can conclude that all the controllers cannot perform well without road information. The road-related features are of crucial importance to train end-to-end controllers. 
The roadside-related features provide the controller with a good generalizability for various scenarios, and therefore should also be included in training data. 
Sky-related features are of least importance and therefore can be excluded to improve the speed of training an end-to-end controller.
Though this work analyzes three specific categories of features and evaluates the effects of each category, the main contribution of this work is proposing a framework for feature analysis and selection, which can be generally utilized for reducing the computational cost of training deep learning-based controllers of autonomous vehicles.

In this paper, we manually classify the features of images for the purpose of clarify and accuracy.
In the future work, methods like fully CNN can be used to automatically classify features, which is advantageous when more features are analyzed. In addition, we only use one human driver to collect data under the constant velocity of 60 km/h.
In the future work, we will use more drivers operating the car under different vehicle velocities. 
Besides, we have shown that the road-related features are the most important features in this work. 
However, it is still not known which part of the road is more important. Based on our driving experience, if we drive at a high speed, we always look further into the lane; when driving at a low speed, we care more about the traffic situation nearby. 
In the future work, we will conduct the road-related feature analysis for different driving conditions. Real driving data sets will also be used in the future research.




\bibliographystyle{IEEEtran}
\bibliography{mybibtex}

\end{document}